\documentclass[letterpaper]{article} % DO NOT CHANGE THIS
\usepackage{aaai25}
\usepackage{times}  % DO NOT CHANGE THIS
\usepackage{helvet}  % DO NOT CHANGE THIS
\usepackage{courier}  % DO NOT CHANGE THIS
\usepackage[hyphens]{url}  % DO NOT CHANGE THIS
\usepackage{graphicx} % DO NOT CHANGE THIS
\usepackage{natbib}  % DO NOT CHANGE THIS AND DO NOT ADD ANY OPTIONS TO IT
\usepackage{caption} % DO NOT CHANGE THIS AND DO NOT ADD ANY OPTIONS TO IT
\usepackage{algorithm}
\usepackage{algorithmic}

\usepackage{enumitem}
\usepackage{amsmath}

\usepackage{newfloat}
\usepackage{listings}
\usepackage{multirow}
\usepackage{booktabs}

\DeclareCaptionStyle{ruled}{labelfont=normalfont,labelsep=colon,strut=off} % DO NOT CHANGE THIS
\floatstyle{ruled}
\newfloat{listing}{tb}{lst}{}
\floatname{listing}{Listing}
\title{Can LLMs Beat Humans in Debating? \\ A Dynamic Multi-agent Framework for Competitive Debate}
\author{
    Yiqun~Zhang\textsuperscript{\rm 1}, Xiaocui~Yang\textsuperscript{\rm 1}, Shi~Feng\textsuperscript{\rm 1}\thanks{Corresponding author.}, Daling~Wang\textsuperscript{\rm 1}, Yifei~Zhang\textsuperscript{\rm 1}, Kaisong~Song\textsuperscript{\rm 2}\\
}
\affiliations{
    \textsuperscript{\rm 1}School of Computer Science and Engineering, Northeastern University, Shenyang, China\\
    \textsuperscript{\rm 2}Alibaba Group, Hangzhou, China
}

\usepackage{bibentry}
\begin{document}

\maketitle

\begin{abstract}
% Competitive debate is a comprehensive and complex computational argumentation task. Large Language Models (LLMs) encounter hallucinations and lack competitiveness in this task. 
Competitive debate is a complex task of computational argumentation. Large Language Models (LLMs) suffer from hallucinations and lack competitiveness in this field.
To address these challenges, we introduce Agent for Debate (Agent4Debate), a dynamic multi-agent framework based on LLMs designed to enhance their capabilities in competitive debate. Drawing inspiration from human behavior in debate preparation and execution, Agent4Debate employs a collaborative architecture where four specialized agents, involving Searcher, Analyzer, Writer, and Reviewer, dynamically interact and cooperate. These agents work throughout the debate process, covering multiple stages from initial research and argument formulation to rebuttal and summary. To comprehensively evaluate framework performance, we construct the Competitive Debate Arena, comprising 66 carefully selected Chinese debate motions. We recruit ten experienced human debaters and collect records of 200 debates involving Agent4Debate, baseline models, and humans. The evaluation employs the Debatrix automatic scoring system and professional human reviewers based on the established Debatrix-Elo and Human-Elo ranking. Experimental results indicate that the state-of-the-art Agent4Debate exhibits capabilities comparable to those of humans. Furthermore, ablation studies demonstrate the effectiveness of each component in the agent structure.
\end{abstract}

\begin{links}
    \link{Code}{https://github.com/ZhangYiqun018/agent-for-debate}
    % \link{Code}{https://anonymous.4open.science/r/agent-for-debate-CB62}
    % \link{Datasets}{https://aaai.org/example/datasets}
    % \link{Extended version}{https://aaai.org/example/extended-version}
\end{links}

\section{Introduction}

Competitive debate, as a structured and competitive form of communication \cite{debatehistory1, debatehistory2}, plays a crucial role in fields such as education, law, and politics. It challenges the comprehensive ability of participants, including logical thinking, expression skills, rapid analysis, argument construction, and rebuttal techniques, ultimately aiming to persuade a third party. With the advancement of artificial intelligence technologies, computational argumentation has emerged, and it is dedicated to simulating and understanding human argumentation processes through computational methods \cite{Atkinson2017TowardsAA, eger2017neural}. However, existing research is largely confined to specific tasks on particular datasets, such as argument mining \cite{lawrence-reed-2019-argument}, argument quality assessment \cite{wachsmuth2017computational}, and argument generation \cite{li2021document}. While these methods excel at specific tasks, they struggle to handle the complexity of competitive debate characterized by its openness, intense competition, and the need for decision-making and comprehensive skills.

\begin{figure}[!t]
    \centering
    \includegraphics[width=\linewidth]{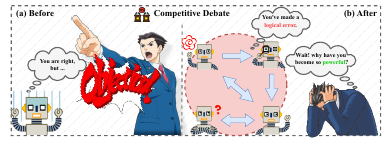}
    \caption{Before and After: Agent4Debate's impact on LLMs competitive debating skills.}
    \label{fig: intro}
\end{figure}

In recent years, Large Language Models (LLMs) \cite{openai2023gpt, touvron2023llama} have demonstrated remarkable capabilities in various natural language processing tasks, offering new possibilities for constructing high-performance debate systems.
Competitive debate, characterized by multi-turn \textbf{document-level} text generation with inter-turn logical dependencies, presents a unique challenge for LLMs, particularly in two significant areas.
First, LLMs often face hallucination problems \cite{ji2023survey}, where models may generate plausible information that is inaccurate or fabricated. Second, due to limitations in safety alignment during training \cite{ouyang2022training} and constraints in handling long contexts \cite{liu2024lost}, models often need to improve in adversarial and sustained debate scenarios (shown in Figure \ref{fig: intro}), struggling to maintain competitiveness and argumentative consistency.

To address these challenges, we propose a multi-agent framework based on LLMs, \textbf{Agent for Debate} (\textbf{Agent4Debate}). Agent4Debate features a dynamic, multi-agent collaborative architecture, leveraging the cooperation of multiple specialized LLMs to enable the framework to participate in multi-stage competitive debates. Inspired by human debate preparation processes, our framework incorporates four key agents, including Searcher, Analyzer, Writer, and Reviewer.
% Our framework demonstrates performance comparable to the human-level in competitive debate. 
To comprehensively evaluate the competitive debate capabilities of Agent4Debate, we establish the \textbf{Competitive Debate Arena}, employing an Elo ranking system widely used in competitive sports, ensuring fairness and scalability.
This arena comprises 66 carefully selected Chinese debate motions, covering three categories \cite{abell2018resolutions}, such as \textit{Policy}, \textit{Value}, and \textit{Fact}, thoroughly testing the performance of participants across different types of debates. Participants include Agent4Debate with different foundation models, two baselines, and ten experienced \textbf{human} debaters. All participants engage in pairwise matches, with each debate assessed through two independent evaluation methods, including an automatic debate judging system based on the Debatrix \cite{liang2024debatrix} metrics, and an expert judging system consisting of three professional human reviewers. 
Based on these two sets of independent evaluation results, we construct two separate Elo \cite{elo1967proposed,zheng2023judging} ranking lists, providing a multi-faceted quantitative assessment of participants' performance across various debate motions. The experimental results from \textbf{the arena} demonstrate that Agent4Debate can achieve human-level performance in various types of competitive debates.
% as evidenced by Debatrix and human judgments. 

In conclusion, the main contributions of this work are as follows:

\begin{itemize}[nolistsep, noitemsep]
\item We propose the Agent4Debate, which enhances the performance of LLMs in competitive debates through dynamic multi-agent collaboration. This framework mimics human debate team interactions, with agents adapting roles and strategies. Specifically, it employs the Searcher for information gathering, the Analyzer for strategic assessment, the Writer for argument formulation, and the Reviewer for critical evaluation.
\item We construct the Competitive Debate Arena, a public resource comprising 66 Chinese debate motions and 200 debate matches across \textit{Policy}, \textit{Value}, and \textit{Fact} categories. Human debaters are incorporated, and we establish Debatrix-Elo and Human-Elo rankings using Debatrix metrics and professional human judges, respectively.
\item Our experimental results indicate that Agent4Debate's performance in competitive debates is comparable to that of humans. Ablation studies validate the effectiveness of each component.
\end{itemize}

\section{Related Work}

\subsection{Computational Argumentation}
Argumentation research has deep historical roots \cite{Walton2008ArgumentationS}, with its core objective being to achieve persuasion through logical reasoning and promote consensus among parties. In recent years, computational argumentation has emerged as an increasingly important field in natural language processing, with its main research directions encompassing argument mining \cite{lawrence-reed-2019-argument, chen2024exploringpotentiallargelanguage}, argument generation \cite{hua2019argumentgenerationretrievalplanning}, argument persuasiveness \cite{carlile-etal-2018-give}, and argument quality assessment \cite{wachsmuth-etal-2017-computational, liang2024debatrix, wachsmuth2024argumentqualityassessmentage}. 
Project Debater \cite{slonim2021autonomous}, a debate system that integrates multiple modules, relies on retrieval-based methods rather than generative approaches for its argumentation. 
With the rise of Large Language Models research utilizing adversarial methods such as debate to enhance model capabilities \cite{du2023improvingfactualityreasoninglanguage, chang2024socrasynthmultillmreasoningconditional} has gradually attracted academic attention. Against this backdrop, our study focuses on competitive debate, a complex computational argumentation task that integrates multiple sub-tasks.

\subsection{LLM-based Agents}
LLMs, such as ChatGPT \citep{openai2023gpt}, LLaMA \citep{touvron2023llama, touvron2023llama2openfoundation}, demonstrate powerful capabilities in instruction following and reasoning tasks.
Harnessing these advanced capabilities, researchers have developed LLM-based agents, which mark a significant step forward in the field. These agents leverage the language understanding and generation abilities of models for more sophisticated tasks like multi-step reasoning and interactive problem-solving, as shown in recent studies \cite{wangSurveyLargeLanguage2023a,liCAMELCommunicativeAgents2023}.
They have various uses across different domains, such as software engineering \cite{qianCommunicativeAgentsSoftware2023} and scientific inquiry \cite{boikoEmergentAutonomousScientific2023}, highlighting their versatility. These agents can imitate complex human actions, partake in social interactions \cite{parkGenerativeAgentsInteractive2023, tuCharacterChatLearningConversational2023}, and replicate intricate scenarios like elections \cite{argyleOutOneMany2022}, debates \cite{wangApolloOracleRetrievalAugmented2023, du2023improvingfactualityreasoninglanguage}, and consumer patterns \cite{wangWhenLargeLanguage2023}, illustrating their capacity to emulate human social dynamics. 
While these agents demonstrate impressive capabilities in emulating human social dynamics, current research predominantly explores collaborative scenarios. However, competitive settings, though equally crucial in human interactions, remain comparatively underexplored.

\section{Task Definition}
% \begin{figure}[!b]
%     \centering
%     \includegraphics{latex/figures/competitive_debate.pdf}
%     \caption{Competitive debate structure.}
%     \label{fig: debate_format}
% \end{figure}
Competitive debate is a structured multi-turn interactive task. Each turn of statement can be regarded as a \textbf{document-level} text generation task, with a temporal and logical progression relationship between multiple turns. 
A typical debate has two opposing sides: the \textit{Pro side} and the \textit{Con side}.
We represent the competitive debate as a sequence: 
\begin{equation}
  D = \{ (s_1, r_1), (s_2, r_2), \cdots, (s_n, r_n)\}  
\end{equation}
where $(s_i, r_i)$ denotes the $i$-th statement and its corresponding role, $s_i$ is the statement, and $r_i \in \{\textit{Pro}, \textit{Con}\}$ represents the role of speaker. Each statement can be defined as:
\begin{equation}
  s_i = \mathcal{G}(m, r_i, D_{i-1}) 
\end{equation}
where $m$ is the motion of debate, $D_{i-1}$ represents the history of the first $i-1$ statements,  and $\mathcal{G}(\cdot)$ is the generation function that produces each statement.

Typical competitive debate structure usually comprises three distinct stages, namely \textit{constructive arguments}, \textit{rebuttals}, and \textit{summary statements}. 
% The format is illustrated in Figure \ref{fig: debate_format}.
To ensure fairness and simulate actual competitive debate conditions \cite{wilson_debate_formats}, we establish specific rules for each stage:

\begin{figure*}[!t]
    \centering
    \includegraphics{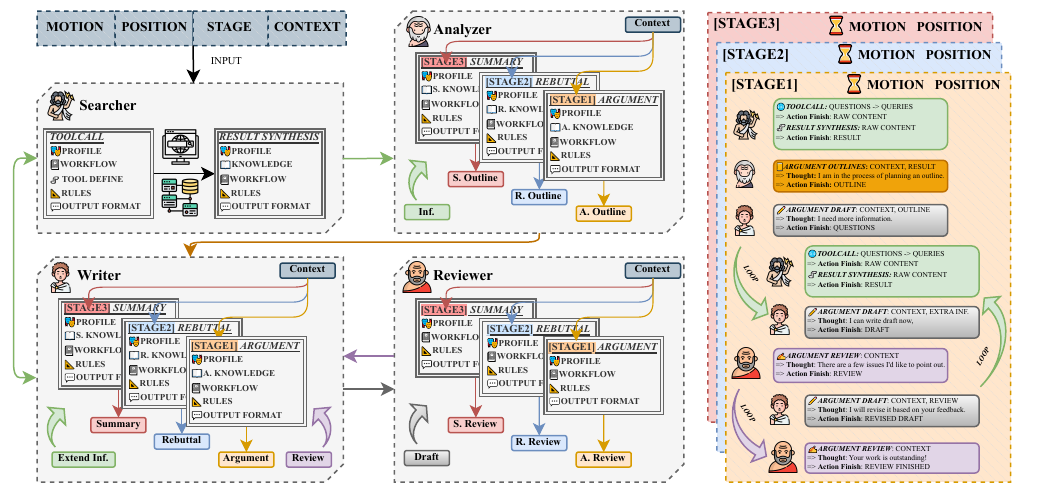}
    \caption{Agent for Debate (Agent4Debate) Workflow: A dynamic framework simulating human debate team collaboration. From searching to reviewing, it showcases how four key roles (Searcher, Analyzer, Writer, Reviewer) interact and work iteratively. The right side illustrates the cyclical process from information gathering to argument formation using \textit{Stage 1} as an example, highlighting the framework's multi-steps progression and recursive refinement.}
    \label{fig: framework}
\end{figure*}

\begin{itemize}[nolistsep, noitemsep]
    \item In \textit{Stage 1} (Constructive Arguments), both sides work independently, with the Con side unable to view the Pro's constructive argument, ensuring initial viewpoints are uninfluenced. 
    \item \textit{Stages 2} and \textit{3} (Rebuttal and Summary) employ a progressive disclosure mechanism, where participants access all previous content to construct targeted statements.
    \item We alternate the sequence across stages to balance the advantages of speaking order. The Pro side speaks first in \textit{Stage 2}, while the Con side leads in \textit{Stage 3}.
\end{itemize}

\section{Agent for Debate}

To address the challenges of hallucination and the difficulties in maintaining competitiveness and argumentative consistency in sustained debate scenarios, we propose the Agent for Debate (Agent4Debate) framework to enable LLMs to participate in competitive debates, as shown in Figure \ref{fig: framework}.
This framework dynamically simulates human debate preparation through dialogue-based collaboration \cite{wu2023autogenenablingnextgenllm} among four LLM-based agents, each mirroring key roles in a human debate team. The \textbf{Searcher} acts as a research assistant, gathering relevant information, while the \textbf{Analyzer} functions like an executive coach, strategizing and analyzing arguments. The \textbf{Writer} performs as a debater, crafting and articulating arguments, and the \textbf{Reviewer} serves as a debate coach, providing feedback and quality control. These agents interact flexibly throughout the debate process, adapting their roles and contributions based on the current stage and needs, much like a well-coordinated human debate team.

The collaboration in Agent4Debate is not just a simple sequence of steps, but rather a dynamic interaction between multiple agents, based on the debate stage and context. 
% For instance, the Writer can request the Searcher to find specific information during the drafting process. 
All the agents are equipped with customized prompts for different debate stages, enabling them to better adapt to and execute the specific tasks of the current stage.
In the following sections, we introduce the functions of each agent in detail. 

\subsection{Searcher Agent}
The Searcher is a tool agent in the Agent4Debate framework, designed to effectively mitigate hallucination issues and address information timeliness problems that LLMs may encounter during debates. It achieves this by accessing and organizing information from external knowledge bases. The workflow of Searcher primarily involves decomposing search questions into more refined queries, then utilizing external tools (such as search engines or specialized knowledge bases) to retrieve relevant information, and finally systematically compiling and organizing the obtained answers.
The information compiled by the Searcher forms a motion knowledge base, which is fixed and accessible to all agents for reference throughout the entire debate process. This approach ensures consistency and reliability of the information used in the debate.
Note that, the Searcher plays different roles at various stages of the debate. In \textit{Stage 1}, the Searcher uses the motion as the search question for information gathering. However, in \textit{Stage 2} and \textit{Stage 3}, the Searcher switches to a passive mode, waiting for specific instructions from the Writer before conducting targeted searches.

\subsection{Analyzer Agent}

The Analyzer is a core agent in the Agent4Debate framework, integrating real-time information from the debate and providing structured guidance for subsequent content output. Its primary function is to systematically analyze and plan the debate content based on the given motion, current stage, and historical context, thus bridging different phases of the debate.
The workflow of Analyzer primarily involves breaking down the debate content step-by-step, drafting detailed outlines, and providing targeted strategic advice to other agents. This approach ensures coherence in debate reasoning and comprehensiveness in argumentation.
Notably, the Analyzer plays different roles at various stages:

\begin{itemize}[nolistsep, noitemsep]
    \item In \textit{Stage 1}, the Analyzer receives the debate topic and compiled materials from the Searcher. It then summarizes the motion and formulates definitions, judgment criteria, main arguments, and supporting evidence from its own perspective.
    \item In \textit{Stage 2}, the Analyzer analyzes all content from previous phases, summarizing the differences in viewpoints between both sides, such as the opponent's definitions and judgment criteria. It then suggests rebuttal techniques that can be used to address these differences.
    \item In \textit{Stage 3}, in addition to continuing to summarize points of disagreement and provide rebuttal techniques, the Analyzer also offers suggestions from a value-based perspective, further enhancing the depth and persuasiveness of the debate.
\end{itemize}

\subsection{Writer Agent}
The Writer is the executive agent in the Agent4Debate framework, responsible for transforming analysis and planning into actual debate content. Its primary function is to compose complete debate drafts based on the instructions and outlines provided by the Analyzer and to revise these drafts according to feedback from the Reviewer, ensuring the quality and persuasiveness of the debate.
Workflow of the Writer primarily encompasses the following aspects:
\begin{itemize}[nolistsep, noitemsep]
    \item \textbf{Content Creation:} Based on the outline provided by the Analyzer, the Writer expands it into a detailed debate script, ensuring the logic of arguments and the sufficiency of supporting evidence.
    \item \textbf{Revision and Refinement:} Upon receiving modification suggestions from the Reviewer, the Writer makes corresponding adjustments and optimizations to the script to enhance its overall quality.
    \item \textbf{Resource Assessment:} The Writer evaluates whether the information in the current knowledge base is sufficient to support the requirements of the outline and script revisions. If information is found to need to be improved, the Writer proactively initiates requests to the Searcher, clearly specifying the additional materials needed.
\end{itemize}

\subsection{Reviewer Agent}
The Reviewer is the quality control agent in the Agent4Debate framework, responsible for reviewing and evaluating debate scripts generated by the Writer. Its primary function is to provide targeted modification suggestions based on the current debate stage and historical context, ensuring the debate content's quality, logic, and persuasiveness. 
The Reviewer's workflow focuses on different aspects at various stages of the debate:

\begin{itemize}[nolistsep, noitemsep]
    \item In \textit{Stage 1}, the Reviewer primarily concentrates on the completeness of the argument structure, the comprehensiveness of content (including definitions, criteria, and main points), the sufficiency of supporting evidence, and the fluency of expression.

    \item In \textit{Stage 2}, building upon the previous stage, the Reviewer additionally focuses on the appropriate application of rebuttal techniques and ensures that rebuttals to the opponent's arguments do not lead to self-contradiction in one's stance.

    \item In \textit{Stage 3}, besides addressing the content from the previous two stages, the Reviewer also assesses the depth of the debate content and makes a judgment based on the context, providing detailed reasons for this assessment.
\end{itemize}

The Reviewer maintains argumentative coherence by continuously assessing consistency with previously presented information across all debate stages. This process involves providing feedback and modification suggestions to the Writer, and facilitating targeted revisions. The review-revision cycle persists iteratively until the script meets the Reviewer's quality standards.

\section{Experimental Setup}

\subsection{Experimental Subjects}
Our experimental design involves three types of participants, including the \textbf{baseline} framework, \textbf{Agent4Debate} based on different LLMs,  and \textbf{human participants}.
For all models, we set temperature to 0.2 and Top \textit{P} to 0.75, with no other parameters adjusted.

\subsubsection{Baseline}

We adopt the benchmark framework of AI-Debater 2024 competition\footnote{http://www.fudan-disc.com/sharedtask/AIDebater24}, incorporating Tavily\footnote{https://tavily.com} as the search engine and stage-specific prompts. We uses Claude-3.5-sonnet and Deepseek-Chat \cite{bi2024deepseek} as the foundation models.

\subsubsection{Agent4Debate}

To comprehensively evaluate the generalization capability of Agent4Debate and conduct more in-depth comparative experiments, we select a variety of advanced LLMs as the foundation for Agent4Debate. These models include Claude-3.5-sonnet, GPT-4o \cite{openai2023gpt}, and Gemini-1.5-Pro/Flash \cite{reid2024gemini}, all of which have demonstrated excellent performance in various evaluations \cite{zheng2023judging}. 
Considering that our study focuses on Chinese competitive debate, we specifically incorporate several LLMs that excel in Chinese language processing, including Qwen2-72b-Instruct \cite{yang2024qwen2}, Deepseek-Chat , and GLM-4-Air. 
Switching models in Agent4Debate experiments updates all components accordingly. In all experiments, the Searcher used Tavily as the search engine.
% It is important to emphasize that when we conduct experiments with different models based on Agent4Debate, all components within the system are correspondingly switched to the newly selected model.

\subsubsection{Humans}

We recruit ten experienced debaters for our experiment to validate the performance of Agent4Debate against humans in competitive debate. Each debater has with 2-4 years of debate team training and at least one year of Chinese competitive debate experience.  They are informed that they will be debating against artificial intelligence and are given 2 days of preparation time for each motion. 
To ensure effective communication, we use the Whisper model \cite{radford2023robust} to transcribe human speeches into text while the human debaters read the model's output directly. This design ensures accurate information transfer and provides human debaters ample time for reflection and response.
\textit{These debaters participate only in the debates, not in other research activities.}

\subsection{Metrics}

\subsubsection{Debatrix}
\label{sec: debatrix}
Debatrix \cite{liang2024debatrix} is a multi-turn debate evaluation method based on LLMs. It comprehensively assesses debates by considering the chronological order of statements and evaluating them along three dimensions, each described in natural language: \textbf{Argument (A)}, \textbf{Source (S)}, and \textbf{Language (L)}. These natural language evaluations are then integrated to form an \textbf{Overall (O)} assessment, ultimately determining the winner. In our implementation, we convert each dimension's descriptive result into a ternary outcome (win, lose, or tie). This evaluation approach is particularly well-suited for our multi-turn, document-level competitive debate scenarios.
In our experiments, we employ GPT-4o-mini as the foundational model for Debatrix. To ensure the reliability of the assessment, we conduct three independent evaluations using Debatrix for each debate, ultimately deriving the final scores.

\subsubsection{Human}
We invite three experienced Chinese competitive debate judges to participate in this study. Each judge possesses 3-5 years of experience in Chinese competitive debates and has coached university debate teams.
The judges independently assess each debate, casting a vote for win, lose, or tie, with the outcome determined by majority rule. To maintain impartiality, judges are only informed that \textbf{both sides have an equal burden of proof} without receiving any additional context. 
\textit{It is important to note that all judges are external to the research development process and do not have backgrounds in computer science, thereby minimizing potential biases.}

\subsection{Competitive Debate Arena}

To comprehensively assess the abilities of Agent4Debate, Baseline, and Humans in competitive debate, we establish the \textbf{Competitive Debate Arena}. This arena is designed to provide a fair and extensible evaluation environment, covering various types of debate motions and assessment methods. We carefully select 66 debate motions from major Chinese debate competitions over the past decade, including \textbf{Chinese Debate World Cup}, \textbf{The World Mandarin Debating Championship}, and \textbf{International Chinese Debating Competition}. These motions cover three main categories \cite{abell2018resolutions}, including \textit{Value}, \textit{Fact}, and \textit{Policy}. Fact makes statements or comparisons about testable aspects of the natural world, Value assigns value or judgment to certain things or concepts, while Policy typically suggests action plans through proposed changes.

In terms of evaluation methods, we adopt two independent review approaches, where one uses the Debatrix based on LLMs for assessment, and the other involves judgments by experienced human reviewers. These review methods are completely independent, each producing separate results. Based on these review methods, we construct two ranking systems, including \textbf{Debatrix-Elo} and \textbf{Human-Elo}.
To build these ranking systems, we draw inspiration from the Chatbot Arena \cite{zheng2023judging} approach and adopt an improved version of the Bradley-Terry (BT) model \cite{hunter2004mm, rafailov2024directpreferenceoptimizationlanguage} to calculate Elo scores. The traditional BT model uses the following formula to calculate the probability of Participant \textit{A} winning over Participant \textit{B}:
\begin{equation}
    P(A > B) = \frac{e^{\gamma_A}}{e^{\gamma_A} + e^{\gamma_B}}
\end{equation}
where $\gamma_A$ and $\gamma_B$ represent the ability parameters of \textit{A} and \textit{B}, respectively.

However, considering that our review system (whether Debatrix or human reviewers) independently provides three scores, we improve the traditional model by introducing a weight function based on score differences:
\begin{equation}
    w_i = \frac{1}{1 + e^{-|\text{score}_{A_i} - \text{score}_{B_i}|}}
\end{equation}
where $\text{score}\in [0, 3]$. This weight function adjusts the importance of each match in the final ranking, making the ranking calculation more precise. Based on this weight function, our likelihood function becomes:

\begin{equation}
\label{eq: bt_model}
    \mathcal{L} = \prod_{i=1}^{n} P(A_i > B_i)^{w_i}
\end{equation}
By maximizing this likelihood function, we can obtain more accurate ability parameter estimates, thus constructing a more precise ranking system.

Our improved Elo system not only effectively reflects participants' overall performance in multiple matchups but also allows for more nuanced adjustments based on the specifics of each match. Using two independent review methods and ranking systems, we can better understand the performance of participants and compare potential differences between Debatrix and human reviews. Furthermore, this Elo system is scalable, efficiently incorporating new frameworks or models for ongoing comparative analysis.

\section{Experimental Results}

\subsection{Agent4Debate vs. Baseline}
We conduct a comparative performance evaluation of Agent4Debate against the baselines. Each framework participates in 20 debate matches, including five different motions. 
To ensure fairness, the number of times each framework argued for the Pro and Con sides is balanced. Debatrix is employed as the evaluation criteria. 
Debatrix scoring is applied three times for each debate, with 1 point awarded for each win in the dimensions of \textbf{Argument (A)}, \textbf{Language (L)}, \textbf{Source (S)}, and \textbf{Overall (O)} performance. In the case of a tie, both sides are awarded 0.5 points.

\begin{table}[!htbp]
\centering
\small
\setlength{\tabcolsep}{1mm}
\begin{tabular}{cccccccc}
\toprule
\multirow{2}{*}[-0.5ex]{\textbf{Model}} & \multirow{2}{*}[-0.5ex]{\textbf{Framework}} & \multicolumn{4}{c}{\textbf{Debatrix}}                                      \\ \cmidrule(l){3-6} 
                                &                                     & \textbf{S} & \textbf{L} & \textbf{A} & \textbf{O} \\ \midrule
\multirow{2}{*}{Claude-3.5-sonnet} & Agent4Debate & \textbf{2.83} & \textbf{1.76} & \textbf{2.52} & \textbf{2.62} \\
                                & Baseline                            & 0.17            & 1.24              & 0.48              & 0.38             \\ \midrule
\multirow{2}{*}{Deepseek-Chat}  & Agent4Debate                        & \textbf{2.73}   & \textbf{1.88}     & \textbf{2.31}     & \textbf{2.77}    \\
                                & Baseline                            & 0.27            & 1.12              & 0.69              & 0.23             \\ \bottomrule
\end{tabular}%
\caption{Comparison of Agent4Debate and Baseline.}
\label{tab: baseline}
\end{table}

As shown in Table \ref{tab: baseline}, Agent4Debate enhances the competitive debating performance across both models. For Claude-3.5-sonnet, the Overall score improves from 0.38 to 2.62, while for Deepseek-Chat, it increases from 0.23 to 2.77. These results demonstrate that the Agent4Debate framework effectively enhances the performance of language models of varying scales and types in competitive debate tasks.
Among all metrics, Source shows improvement. This can be attributed to the Searcher Agent and Analyzer Agent within Agent4Debate, which conducts an in-depth analysis of debate motions and systematic organization of materials, utilizing external knowledge more effectively than the simple search approach from baseline. The Language shows relatively modest improvement, reflecting robust generation capabilities of LLMs, leaving limited room for enhancement.

Comparing the results between Claude-3.5-sonnet and Deepseek-Chat, it is observed that Agent4Debate yields more pronounced performance improvements for the more powerful model, particularly in the Argument and Overall metrics. This may be due to more advanced models possessing stronger reasoning abilities and better instruction-following capabilities \cite{kaplan2020scalinglawsneurallanguage}, thus exhibiting superior adaptability to complex frameworks.

\subsection{Ablation Study}
To evaluate the contribution of each agent within Agent4Debate, we conduct a series of ablation studies. The experimental setup remains consistent with the previous comparative experiments. Each ablation configuration engages in 20 debates across five motions, with a balanced distribution of the \textit{Pro} and \textit{Con} sides. The evaluation continues to employ Debatrix, with the scoring method identical to that of the comparative experiments. We do not perform an ablation experiment on the Writer Agent, as it is responsible for the text generation at every stage.
The foundation model for the ablation study is Claude-3.5-sonnet.

\begin{table}[!htbp]
\centering
\small
% \resizebox{\columnwidth}{!}{%
\begin{tabular}{lcccccc}
\toprule
\multirow{2}{*}[-0.5ex]{\textbf{Framework}} & \multicolumn{4}{c}{\textbf{Debatrix}}                                  \\ \cmidrule(l){2-5} 
                           & \textbf{S}        & \textbf{L}      & \textbf{A}      & \textbf{O}       \\ \midrule
Agent4Debate               & \textbf{2.79} & \textbf{1.54} & \textbf{2.01} & \textbf{2.12} \\
w/o Searcher               & 0.21          & 1.46          & 0.99          & 0.88          \\ \midrule
Agent4Debate               & \textbf{1.83} & \textbf{1.50}  & \textbf{1.79} & \textbf{1.76} \\
w/o Analyzer               & 1.17          & \textbf{1.50}  & 1.21          & 1.24          \\ \midrule
Agent4Debate               & \textbf{1.74} & \textbf{1.67} & \textbf{2.13} & \textbf{1.93} \\
w/o Reviewer               & 1.26          & 1.33          & 0.87          & 1.07          \\ \bottomrule
\end{tabular}%
% }
\caption{The results of ablation study. The foundation model for the ablation study is Claude-3.5-sonnet.
}
\label{tab: ablation}
\end{table}

% The experimental results demonstrate that each agent in the Agent4Debate framework contributes to the overall performance. Table \ref{tab: ablation} presents the detailed results of our ablation study, clearly illustrating the impact of removing each agent. Removing any agent leads to a decrease in the Overall score, confirming the necessity of each component.

Table \ref{tab: ablation} presents the detailed results of our ablation study, clearly illustrating the impact of removing each agent. The experimental results demonstrate that each agent in the Agent4Debate framework contributes to the overall performance. When we remove any agent, the Overall score decreases,  confirming the necessity of each component.
Specifically, removing the Analyzer reduces the Overall score from 2.12 to 1.76. Its impact on the Source and Argument metrics is particularly notable, with the Source score dropping from 2.79 to 1.83 and the Argument score from 2.01 to 1.79. This indicates the Analyzer's crucial role in the formulation of material analysis, argument refinement, and rebuttal strategy. The absence of the Searcher results in a dramatic drop in the Source score from 2.79 to 0.21, while the Overall score falls from 2.12 to 0.88. This highlights the importance of appropriately searching and organizing external knowledge to enhance debate performance. The removal of the Reviewer has a smaller impact on overall performance (Overall score decreases from 2.12 to 1.93). However, its primary function of reviewing drafts, suggesting revisions, and improving the output quality of Agent4Debate aligns with the framework's design expectations.

\subsection{Results of Competitive Debate Arena}
We collect records of 200 debate matches (excluding those from comparison experiments and ablation studies), covering 66 debate motions across three categories, including \textit{Fact}, \textit{Value}, and \textit{Policy}. Participants included in Agent4Debate using different foundation models, two baselines, and ten human debaters, all of which are engaged in randomly paired competitions.
Each debate is independently assessed using both the Debatrix and human judges. Utilizing the improved BT model in Eq. \ref{eq: bt_model}, we calculate Elo scores for all 200 matches and sub-Elo scores for each of the three debate categories. 
The experimental results are presented in two independent ranking systems, consisting of Debatrix-Elo (Table \ref{tab: debatrix-elo}) and Human-Elo (Table \ref{tab: human-elo}).
% Models without an asterisk ($*$) indicate the foundation models used by Agent4Debate, while those with an asterisk denote the models used as baselines.

\begin{table}[!htbp]
\centering
\small
\setlength{\tabcolsep}{1.0mm}
\begin{tabular}{lcccccc}
\toprule
\textbf{Model}                    & \textbf{Full} & \textbf{Fact}                            & \textbf{Policy} & \textbf{Value} \\ \midrule
Gemini-1.5-Pro &
\textbf{1034.15} &
1154.93 &
\textbf{1231.98} &
1075.30 \\
Claude-3.5-sonnet &
1032.51 &
\textbf{1159.18} &
1224.19 &
1074.33 \\
Qwen2-72b-Instruct &
1023.31 &
  1130.83 &
1179.62 &
\textbf{1081.75} \\
GPT-4o             & 1022.21          & 1150.14                                  & 1137.49         & 1069.55        \\
Gemini-1.5-Flash & 1012.45          & 1136.21                                  & 1156.50          & 1057.73        \\
GLM-4-Air          & 1011.72          & 1155.07 & 1148.53         & 1048.42        \\
Deepseek-chat      & 1004.00             & 1118.98                                  & 1131.16         & 1054.89        \\
Claude-3.5-sonnet$^*$      & 982.07           & 479.50                                    & 956.21          & 1021.44        \\
Human                             & 978.35           & 1109.73                                  & 515.57          & 953.05         \\ 
Deepseek-Chat$^*$          & 954.34           & 491.13                                   & 478.78          & 983.99         \\ \bottomrule
\end{tabular}%
\caption{Debatrix-Elo Ranking. $^*$ denotes baseline models, unmarked models are Agent4Debate foundation models.}
\label{tab: debatrix-elo}
\end{table}

\begin{table}[!htbp]
\centering
\small
\setlength{\tabcolsep}{1mm}
\begin{tabular}{lcccccc}
\toprule
\textbf{Model}                      & \textbf{Full} & \textbf{Fact} & \textbf{Policy} & \textbf{Value} \\ \midrule
Gemini-1.5-Pro &
\textbf{1040.64} &
\textbf{1110.23} &
\textbf{1104.79} &
\textbf{1048.10} \\
Claude-3.5-sonnet &
1031.15 &
1093.87 &
1104.44 &
1020.05 \\
GPT-4o &
1028.84 &
1086.78 &
1099.63 &
1033.09 \\
Human                               & 1006.46          & 1055.82       & 1030.32         & 1006.57        \\
Gemini-1.5-Flash   & 1000.00             & 1037.45       & 997.66          & 1003.29        \\
Qwen2-72b-Instruct & 999.70            & 1041.10        & 976.16          & 1005.56        \\
Claude-3.5-sonnet$^*$       & 991.38           & 1023.29       & 968.34          & 997.47         \\
GLM-4-Air            & 972.48           & 940.00           & 948.31          & 996.67         \\
Deepseek-chat        & 971.94           & 963.05        & 946.30           & 986.79         \\
Deepseek-Chat$^*$            & 962.61           & 786.44        & 911.33          & 979.29         \\ \bottomrule
\end{tabular}%
\caption{Human-Elo Ranking. $^*$ denotes baseline models, unmarked models are Agent4Debate foundation models.}
\label{tab: human-elo}
\end{table}

Drawing from the experimental results presented in Tables \ref{tab: debatrix-elo} and \ref{tab: human-elo}, we can derive the following insights.
(1) Agent4Debate, especially those using advanced foundation models such as Gemini-1.5-Pro and Claude-3.5-sonnet, demonstrate performance comparable to or surpassing human debaters in Debatrix-Elo and Human-Elo rankings. The top-performing Agent4Debate (Gemini-1.5-Pro) consistently ranks first, scoring 1044.18 in Debatrix-Elo and 1040.64 in Human-Elo. Experimental results indicate that models with more robust reasoning and instruction-following capabilities perform better within the Agent4Debate framework.
(2) In Debatrix-Elo, most models show score variations across the Fact, Policy, and Value categories. In contrast, Human-Elo displays more consistent scores for each model across categories. This disparity may arise because Debatrix considers Source, Language, and Argument dimensions, while human judges likely focus more on logic and rebuttal techniques. 
Debatrix-Elo and Human-Elo show high consistency in model rankings, particularly for top-performing models. However, human performance is ranked differently in the two rankings. In Debatrix-Elo, humans rank 8th with a score of 978.35, while in Human-Elo, they rank 4th with a score of 1006.46. This suggests that Debatrix-Elo may underestimate human performance. This underestimation is partly due to the different evaluation tendencies between Debatrix and human judges, and partly because human speech quality deteriorates when transcribed to text.
(3) In Debatrix-Elo, certain models excel in specific categories. This is due to differences in the argumentation processes for the three types of debate motions: Policy debates typically require extensive evidence to demonstrate policy necessity and effectiveness; Value debates often demand more substantial logical reasoning and expressive skills; Fact debates combine characteristics of both. These distinctions, reflected in Debatrix's multi-dimensional evaluation, yield varying results.

\subsubsection{Agent4Debate vs. Humans}

\begin{table}[!htbp]
\centering
\small
\begin{tabular}{lccccccc}
\toprule
\multirow{2}{*}[-0.5ex]{\textbf{Model}} & \multicolumn{4}{c}{\textbf{Debatrix}}                       & \multirow{2}{*}[-0.5ex]{\textbf{Human}} \\ \cmidrule(lr){2-5}
      & \textbf{S} & \textbf{L} & \textbf{A} & \textbf{O} &      \\ \midrule
Human & 0.52            & 0.30               & 0.6               & 0.42             & 1.22 \\
Agent4Debate                    & \textbf{2.48} & \textbf{2.70} & \textbf{2.40} & \textbf{2.58} & \textbf{1.78}                   \\ \bottomrule
\end{tabular}%
\caption{Comparison of Human and Agent4Debate}
\label{tab: human detailed}
\end{table}

We conduct a separate analysis of 30 debates between Agent4Debate and human debaters. In these debates, to ensure comprehensive experimentation, all foundation models of Agent4Debate participate. The scoring results from the Debatrix system and human judges are presented in Table \ref{tab: human detailed}.

Debatrix for human performance is lower than human judges across three dimensions. This discrepancy may stem from several factors. Regarding Source, human debaters use voice input, which is then transcribed into text. People typically do not directly cite references in oral debates, leading to lower scores. The Language score is the lowest, possibly due to oral expressions often containing verbal tics and informal language, coupled with imperfect voice-to-text transcription accuracy, affecting language quality assessment. The low Argument score may be a cascading effect of the previous two low scores, thus impacting Debatrix's overall understanding and evaluation of human input.
In contrast, human judges employ different criteria when evaluating competitive debates. They usually prioritize core factors such as logical reasoning and debating skills, only considering other aspects when these primary elements are challenging to distinguish. This approach to judgment differs from the Debatrix.

\subsubsection{Consistency}
To analyze the differences between Debatrix and human evaluations, we conduct a consistency analysis.
Consistency is calculated by comparing the result between human and Debatrix, with tie considered consistent outcomes.
Table \ref{tab: consistency} presents the results, showing that internal consistency among human reviewers remains stable across all matches, while the consistency between Debatrix and human reviewers varies when including or excluding human debaters. These findings further corroborate the above observations. 
These findings suggest that while Debatrix shows differences from human reviewers in evaluating debates between humans and models, it still provides valuable insights, particularly in assessing model-to-model debates. In these cases, Debatrix offers multi-faceted analytical results that contribute to our understanding of models' comprehensive capabilities in competitive debates.

\begin{table}[!htbp]
\centering
\small
\setlength{\tabcolsep}{1mm}
\begin{tabular}{lcc}
\toprule
\textbf{Consistency}      & \textbf{Excluding Human Debates} & \textbf{All Debates} \\ \midrule
Debatrix vs. Human & 0.66                             & 0.56                 \\
Among Human        & 0.74                             & 0.73                 \\ \bottomrule
\end{tabular}%
\caption{Consistency between Debatrix and Human Judges}
\label{tab: consistency}
\end{table}

%% arxiv version.
We further analyze the Elo rankings and Agent4Debate in the \textbf{Appendix}.

\section{Conclusion}
We propose a dynamic multi-agent framework, Agent for Debate (Agent4Debate), to enable LLMs to participate in competitive debates. To evaluate the performance of Agent4Debate, we construct the Competitive Debate Arena, comprising 66 classic Chinese debate motions. 
We recruit ten human debaters and collect 200 debate matches involving Agent4Debate, baselines, and human debaters. Using the Debatrix and human judges for evaluation, we propose Debatrix-Elo and Human-Elo rankings. 
Experimental results show that our state-of-the-art Agent4Debate exhibits capabilities comparable to those of humans in competitive debates.
Ablation studies prove the effectiveness of each component in the agent structure.
% \section*{Ethical Statement}

\bigskip

\bibliography{aaai25}

%% arxiv version & supplementary material
\appendix
\section*{Appendix}
\subsection*{Further Analysis of Elo Rankings}

\textbf{Competitive Debate Arena} currently includes only Chinese debate records. This choice is based on the availability of professional Chinese debate judges, ensuring the reliability of our Elo ranking system. However, Agent4Debate is designed to support competitive debates in other languages, including English, using the same structural framework. Expanding to multilingual debates would require only minor adjustments to the language constraints in the \textit{prompts}. Future research may explore the implementation of debates in various languages.

\begin{table}[!htbp]
\centering
\small
\begin{tabular}{lcc}
\toprule
\textbf{Model} & \textbf{Debatrix-CI} & \textbf{Human-CI} \\
\midrule
Gemini-1.5-Pro & + 69/- 18 & + 57/- 14 \\
Claude-3.5-sonnet & + 67/- 18 & + 54/- 14 \\
Qwen2-72b-instruct & + 66/- 20 & + 54/- 15 \\
GPT-4o & + 66/- 20 & + 58/- 14 \\
Gemini-1.5-flash & + 66/- 18 & + 53/- 15 \\
GLM-4-Air & + 67/- 16 & + 48/- 19 \\
Deepseek-chat & + 64/- 22 & + 44/- 19 \\
baseline (Claude-3.5-sonnet) & + 69/- 36 & + 51/- 31 \\
Human & + 64/- 550 & + 50/- 22 \\
baseline (Deepseek-chat) & + 62/- 530 & + 32/- 520 \\
\bottomrule
\end{tabular}
\caption{95\% Confidence Intervals}
\label{tab: confidence_intervals}
\end{table}

Table \ref{tab: confidence_intervals} presents the 95\% confidence intervals (CI) for various models and human performance, as evaluated by both the Debatrix-Elo and Human-Elo rankings. The CIs are expressed as upper and lower bounds relative to the median scores. Based on experimental results, we estimate that new debate models or frameworks can achieve a relatively stable ranking after about 15 debates. This allows for quick assessment of their competitive debate performance. Models with lower win rates show wider CIs (like baseline (Deepseek-Chat)), especially in the lower bound. However, this does not significantly affect the evaluation of their debate performance. The wider CIs mainly reflect the increased uncertainty in precise ranking for these models.

\begin{figure}[!htbp]
    \centering
    \includegraphics[width=\linewidth]{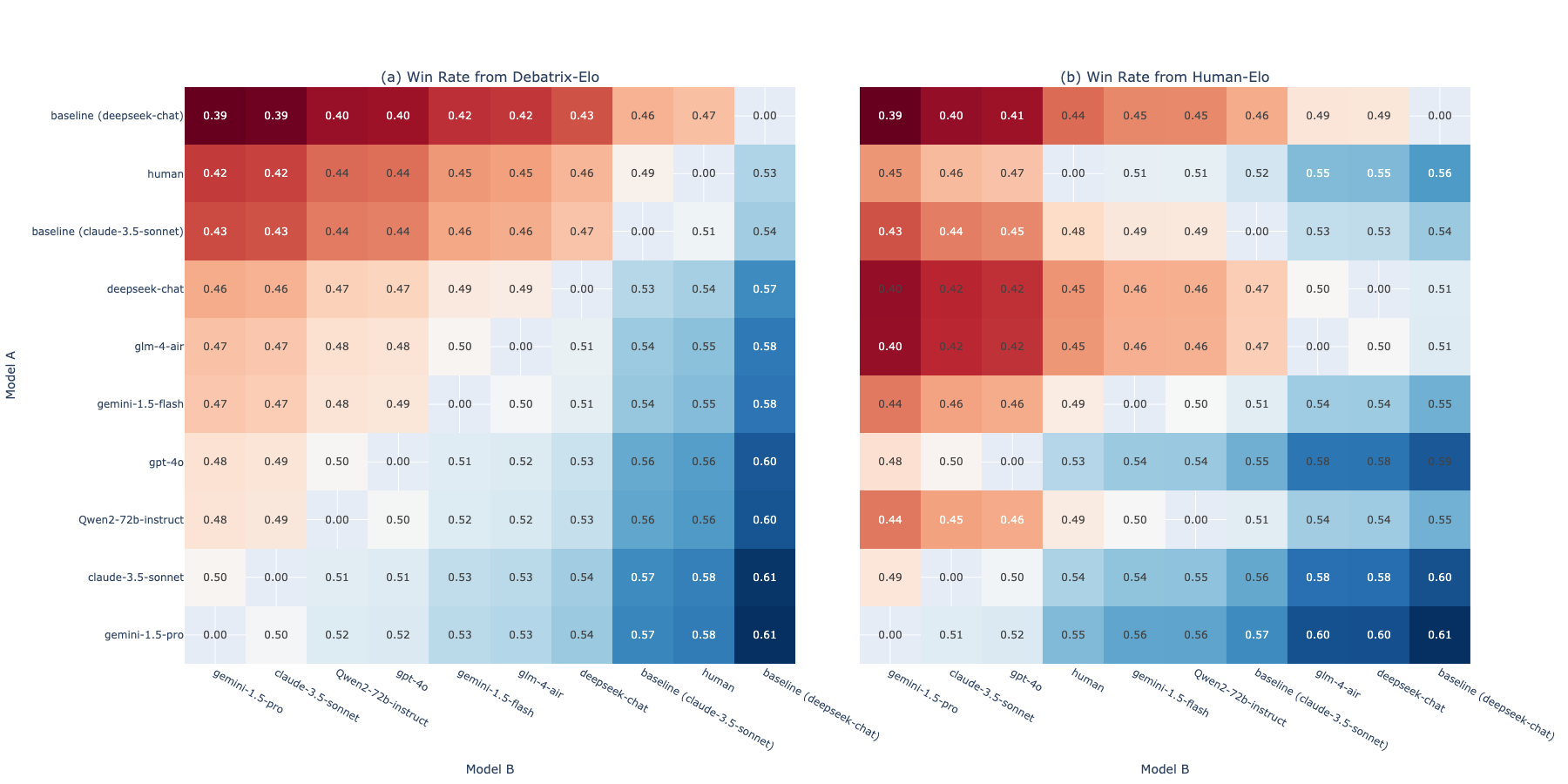}
    \caption{Predicted Win Rates Using Elo Rankings for Model A in A vs. B Battles.}
    \label{fig: rating}
\end{figure}

The Elo system can be used to estimate the winning rate of competitive debates between models.
Figure \ref{fig: rating} presents the win rate calculated using the Debatrix-elo and Human-elo rankings, respectively.

\subsection*{Further Analysis of Agent4Debate}
As shown in Table \ref{tab: agent_calls}, the interaction frequency data reveals distinct patterns across debate stages. The Searcher role shows peak activity during the Argument stage (3.4431), while the Writer's interactions are highest in the Summary stage (3.0932). The Reviewer maintains consistently high engagement throughout all stages. In contrast, the Analyzer exhibits a stable, low interaction frequency (approximately 1.0) across all phases. 
Notably, the Searcher's high activity in the \textit{Stage 1} is due to its proactive information-seeking role, whereas in subsequent stages (Rebuttal and Summary), it assumes a more passive stance, resulting in lower interaction frequencies.
These patterns suggest varied role importance and engagement levels at different debate stages, with the Analyzer serving as an efficient, low-interaction information processor throughout the process.

\begin{table}[!htbp]
\centering
\small
\begin{tabular}{lcccc}
\toprule
\textbf{Stage} & \textbf{Searcher} & \textbf{Analyzer} & \textbf{Writer} & \textbf{Reviewer} \\
\midrule
Overall  & 2.1862 & 1.0015 & 2.7243 & 2.4888 \\
Stage 1 & 3.4431 & 1.0036 & 2.4313 & 2.1682 \\
Stage 2 & 1.3435 & 1.0000 & 2.7939 & 2.5817 \\
Stage 3  & 1.2559 & 1.0000 & 3.0932 & 2.8720 \\
\bottomrule
\end{tabular}
\caption{Interaction Frequencies of Roles Across Debate Stages in the Agent4Debate}
\label{tab: agent_calls}
\end{table}

Table \ref{tab:token_completion} illustrates the average token usage (only calculated completion token) per utterance for each agent. The Searcher, responsible for collecting and organizing information, shows the highest token consumption (2182.90). The Analyzer and Writer exhibit similar token usage (795.06 and 811.94 respectively). The Reviewer, tasked with providing feedback, has the lowest token consumption (415.57), consistent with its role in offering concise critiques. These token consumption patterns align well with the intended functions of each role in our multi-agent framework design, with the Searcher's high token usage emphasizing its critical role in information processing.

\begin{table}[!htbp]
\centering
\small
\begin{tabular}{lcccc}
\toprule
 \textbf{Token} & \textbf{Searcher} & \textbf{Analyzer} & \textbf{Writer} & \textbf{Reviewer} \\
\midrule
Completion & 2182.90 & 795.06 & 811.94 & 415.57 \\
\bottomrule
\end{tabular}
\caption{Token completion counts for different agents}
\label{tab:token_completion}
\end{table}

\subsection*{Prompts}
This section outlines the prompt design principles for Agent4Debate. Due to space constraints, we provide only key examples here, with the complete set of prompts available in the Github Repository\footnote{https://github.com/ZhangYiqun018/agent-for-debate}.

% This section outlines the prompt design principles for Agent4Debate. Due to space constraints, we provide only key examples here, with the complete set of prompts available in the Github Repository\footnote{https://github.com/ZhangYiqun018/agent-for-debate}.

Agent4Debate employs a conversational multi-agent collaborative structure without implicit long-term memory. All information is stored within the dialogue context, accessible to each agent during their turn. The prompt design for each agent consists of five components: \textit{profile}, \textit{knowledge}, \textit{workflow}, \textit{rules}, and \textit{output format}.

The \textit{profile} contains basic agent information. For instance, the Analyzer's \textit{profile} might state: "You are an experienced debate coach tasked with analyzing debate motions, stances, and relevant materials." The knowledge component stores debate-related information and techniques specific to the task of agent. For example, the Writer agent for the constructive argument (\textit{Stage 1}) would have prompts on proof techniques, while \textit{Stage 2} Writer would have prompts on logical fallacies and rebuttal strategies. The \textit{workflow} component outlines the specific steps for task completion, ensuring each agent follows a chain-of-thought (CoT) process. The \textit{rules} component lists guidelines to prevent low-quality responses. Finally, the \textit{output format} defines the structure of the response of agent.

Each agent is guided by prompts with similar structures but varying content across different debate stages, ensuring task completion at each stage.

\subsection*{Case Study}

\begin{figure}[!htbp]
    \centering
    \includegraphics[width=\linewidth]{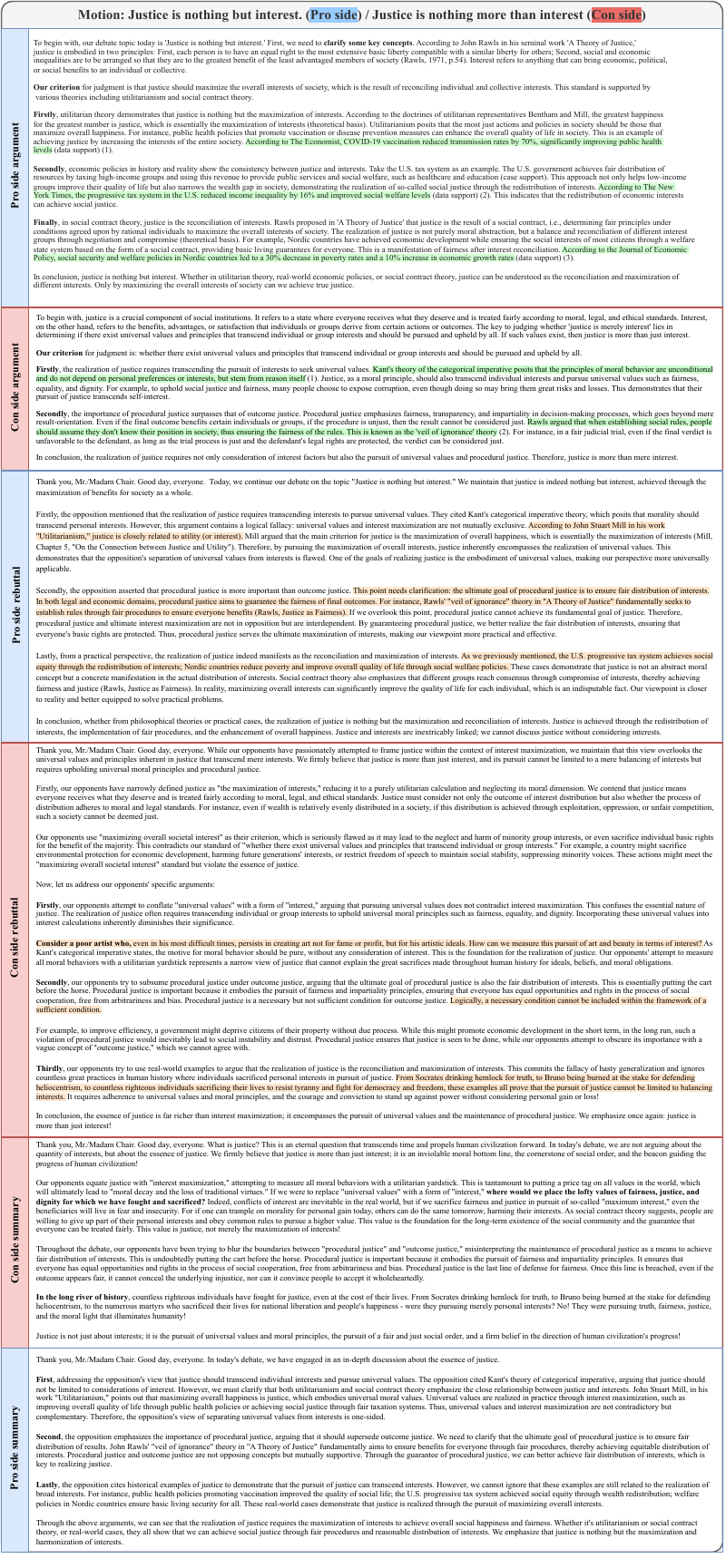}
    \caption{English (translated by Claude-3.5-sonnet from Chinese) case study of the debate motion "Justice is nothing but interest. (Pro side) / Justice is nothing more than interest (Con side)". Pro side is Agent4Debate (GPT-4o), Con side is Agent4Debate (Claude-3.5-sonnet).}
    \label{fig: case1-en}
\end{figure}

\begin{figure}[!htbp]
    \centering
    \includegraphics[width=\linewidth]{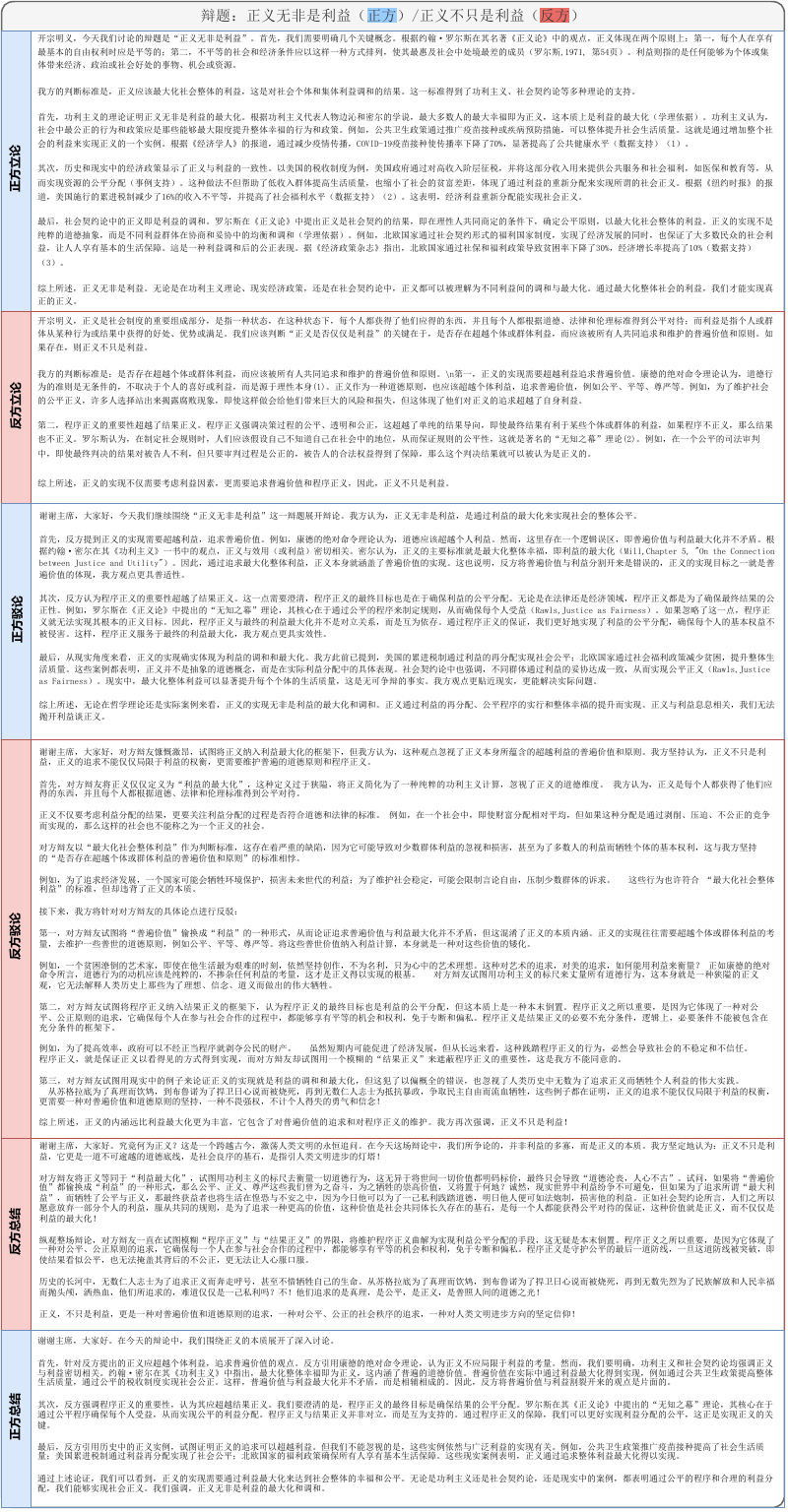}
    \caption{Chinese case study of the debate motion "Justice is nothing but interest. (Pro side) / Justice is nothing more than interest (Con side)". Pro side is Agent4Debate (GPT-4o), Con side is Agent4Debate (Claude-3.5-sonnet).}
    \label{fig: case1-zh}
\end{figure}

\begin{figure}[!htbp]
    \centering
    \includegraphics[width=\linewidth]{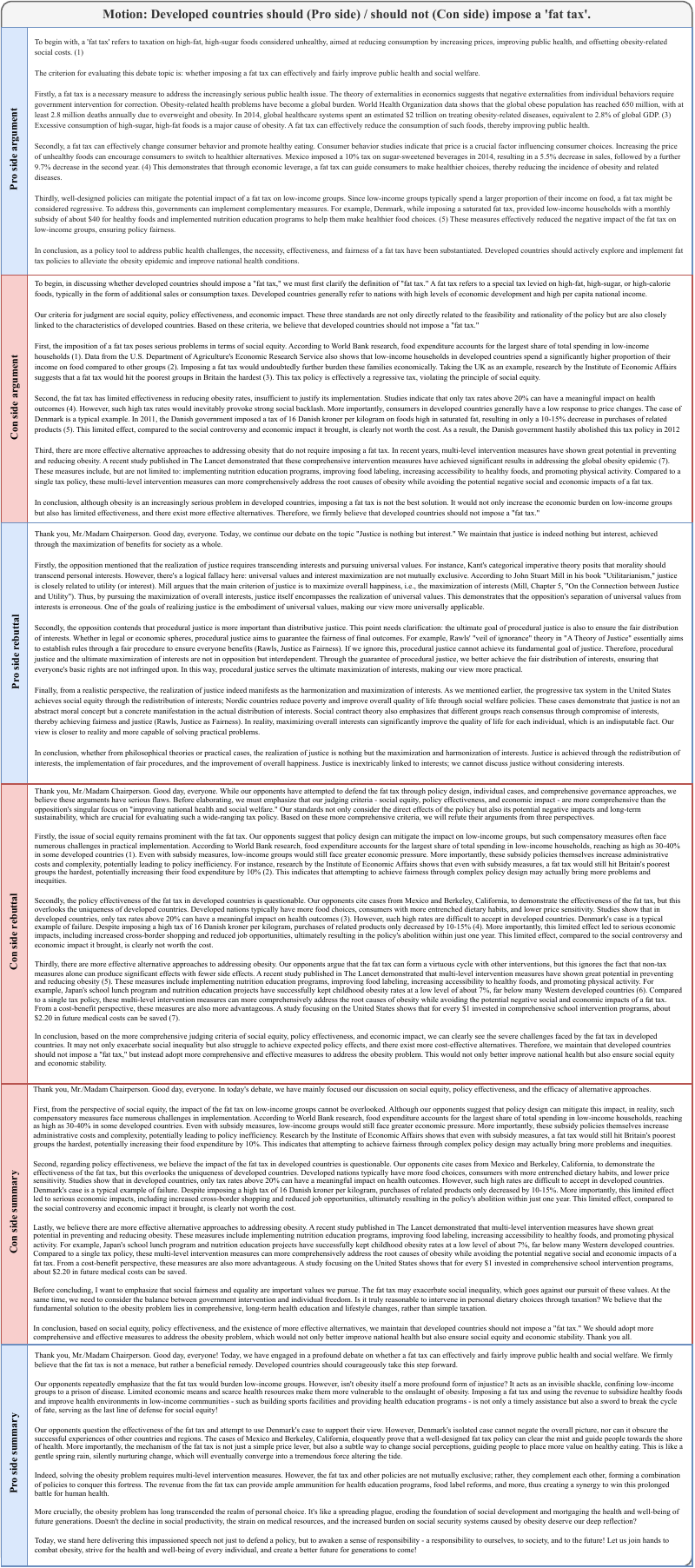}
    \caption{English (translated by Claude-3.5-sonnet from Chinese) case study of the debate motion "Developed countries should (Pro side) / should not (Con side) impose a fat tax.". Pro side is Agent4Debate (Gemini-1.5-Pro), Con side is Agent4Debate (Claude-3.5-sonnet).}
    \label{fig: case2-en}
\end{figure}

\begin{figure}[!htbp]
    \centering
    \includegraphics[width=\linewidth]{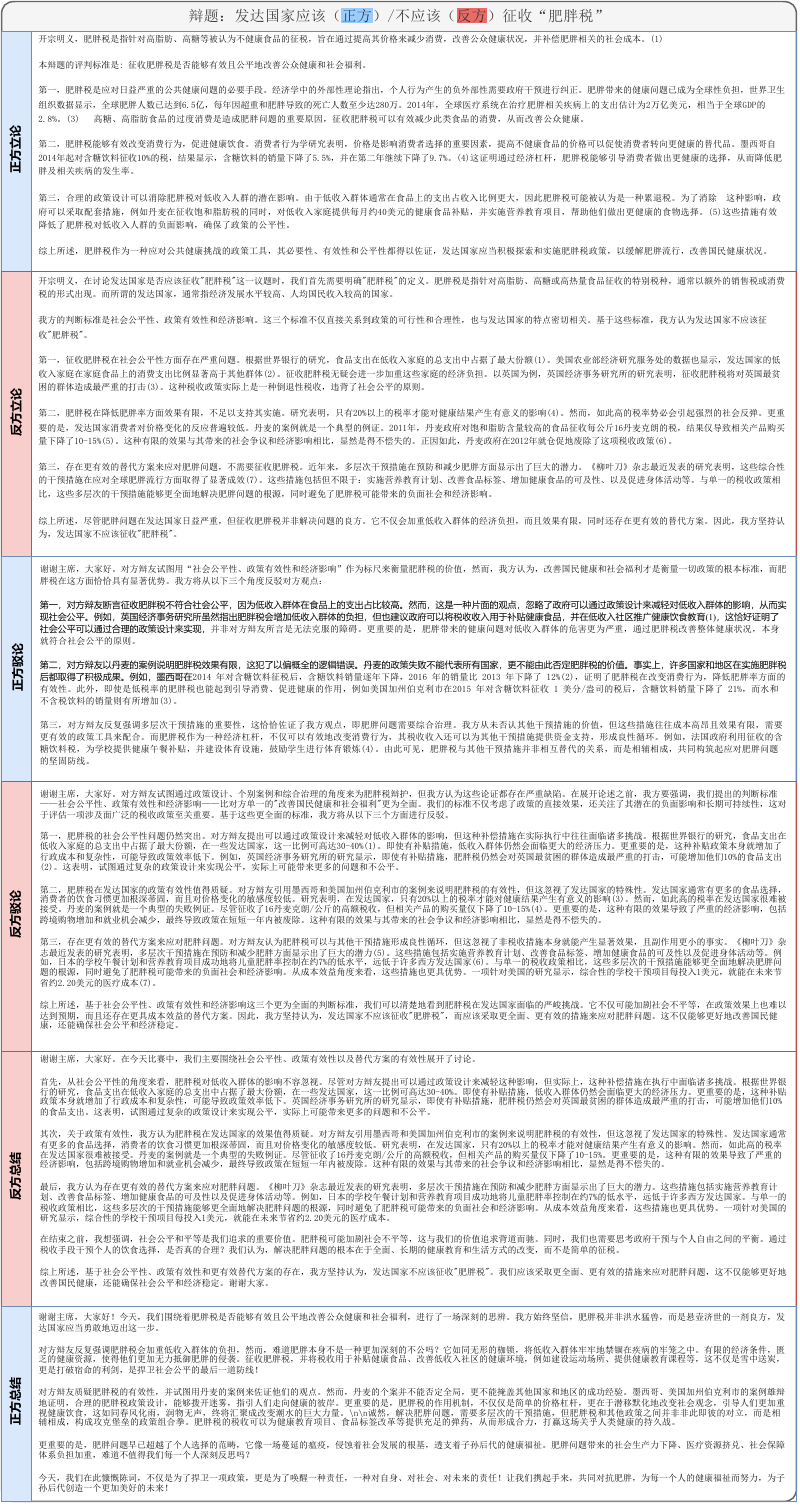}
    \caption{Chinese case study of the debate motion "Developed countries should (Pro side) / should not (Con side) impose a fat tax.". Pro side is Agent4Debate (Gemini-1.5-Pro), Con side is Agent4Debate (Claude-3.5-sonnet).}
    \label{fig: case2-zh}
\end{figure}

Figures \ref{fig: case1-en} to \ref{fig: case2-zh} show case studies of two debate motions: one on the correlation between justice and interest and another on implementing a fat tax in developed countries. Each motion is shown in English and Chinese, with the English versions translated from Chinese using Claude-3.5-sonnet. These case studies demonstrate the application of Agent4Debate using different foundation models, including GPT-4o, Claude-3.5-sonnet, and Gemini-1.5-Pro. Due to space limitations, references for the case studies have been omitted.

\end{document}